\documentclass[11pt,a4paper]{article}
\usepackage[hyperref]{acl2021}
\usepackage{times}
\usepackage{latexsym}

\usepackage{microtype}

\aclfinalcopy 

\usepackage{amsmath,amsfonts,bm}

\def\eqref#1{equation~\ref{#1}}

\def\1{\bm{1}}

\DeclareMathAlphabet{\mathsfit}{\encodingdefault}{\sfdefault}{m}{sl}
\SetMathAlphabet{\mathsfit}{bold}{\encodingdefault}{\sfdefault}{bx}{n}

\usepackage[ruled,vlined, norelsize]{algorithm2e}
\usepackage{booktabs}
\usepackage{multirow}
\usepackage{graphicx}
\usepackage{subfigure}
\usepackage[T1]{fontenc}

\newcommand{\conseq}{\textsc{ConSeq}}
\usepackage{todonotes}

\title{Improving Factual Consistency of Abstractive Summarization via Question Answering}

\author{Feng Nan$^1$ \quad Cicero Nogueira dos Santos$^1$ \quad Henghui Zhu$^1$ \quad Patrick Ng$^1$ \\ \textbf{Kathleen McKeown}$^{1,2}$ \quad \textbf{Ramesh Nallapati}$^1$ \quad \textbf{Dejiao Zhang}$^1$ \\ \quad \textbf{Zhiguo Wang}$^1$ \quad \textbf{Andrew O. Arnold} $^1$ \quad \textbf{Bing Xiang}$^1$ \\ Amazon Web  Services$^1$, Columbia University$^2$\\
	\texttt{\{nanfen, cicnog, henghui, patricng, mckeownk} \\ \texttt{rnallapa, dejiaoz, zhiguow, anarnld, bxiang\}@amazon.com}		
}

\date{}

\begin{document}
\maketitle
\begin{abstract}
	A commonly observed problem with the state-of-the art abstractive summarization models is that the generated summaries can be factually inconsistent with the input documents. The fact that automatic summarization may produce plausible-sounding yet inaccurate summaries is a major concern that limits its wide application. 
In this paper we present an approach to address factual consistency in summarization. 	
	We first propose an efficient automatic evaluation metric to measure factual consistency; 
	next, we propose a novel learning algorithm that maximizes the proposed metric during model training. 
	Through extensive experiments, we confirm that our method is effective in improving factual consistency and even overall quality of the summaries, as judged by both automatic metrics and human evaluation. 
\end{abstract}

\section{Introduction}
Recent advances in neural text generation have led to significant  improvement in the quality of abstractive summarization \cite{radford2019language, gehrmann-etal-2019-generating, lewis2019bart}.
Despite this progress,
there are still many limitations facing neural text summarization \cite{kryscinski-etal-2019-neural},
the most serious of which is the tendency to generate summaries that are not factually consistent with the input document; 
a factually consistent summary only contains statements that can be 
inferred from the source document. Recent studies show that about 30\% of the summaries generated by neural network sequence-to-sequence (seq2seq) models suffer from fact fabrication \cite{AAAI1816121}. 

The standard training approach for seq2seq learning has been maximizing the log likelihood of the target given the input sequences (MLE). It has empirically performed well as a surrogate loss for evaluation metrics such as BLEU and ROUGE. This empirical success can be ascribed to the fact that both BLEU and ROUGE are directly linked to the n-gram overlap between the output and the target sequences, which can be efficiently learned via MLE. In contrast, metrics to capture factual consistency are much more elusive as 
they must take into account the relations among tokens in the context of an entire sequence. The widely used ROUGE score is inadequate to quantify factual consistency \cite{kryscinski-etal-2019-neural}. In fact, the lack of an effective (automatic) metric for factual consistency has been the major hurdle in improving abstractive summarization model training beyond MLE. 
\begin{table}[]
	\centering
	\resizebox{0.5\textwidth}{!}{%
		\scriptsize
		\begin{tabular}{p{0.1\linewidth}p{0.8\linewidth}}
			\toprule
			Input: & ...``Klitschko doesn't have the legs, the power that he used to,'' said Lewis. ``He has a chink in his armour after getting beat by Tyson Fury. Anthony Joshua is now taking that challenge, going after the man.'' ...
			\\  \cmidrule(l){1-2} 
			MLE: & \textcolor{red}{Anthony Joshua has a ``chink in his armour''} ahead of his world heavyweight title bout with Wladimir Klitschko, says former champion Lennox Lewis.
			\\  \cmidrule(l){1-2} 
			\conseq: & Wladimir Klitschko has a ``chink in his armour'' and is no match for British champion Anthony Joshua, says former world heavyweight champion Lennox Lewis.
			\\ \bottomrule
		\end{tabular}%
	}
	\caption{Example summaries from the BART-large finetuned models on test set. Standard 
	MLE training generates a factually inconsistent summary whereas our proposed \conseq\ is  consistent. 
	}
	\label{tab:qualitative_results}
\end{table}
Table \ref{tab:qualitative_results} shows an example of a factually inconsistent summary generated by fine-tuning the BART-large model \cite{lewis2019bart}, which is a transformer based seq2seq model pre-trained on a large corpus with denoising objectives. Standard MLE training produces summaries with factual errors that,
in addition to hallucinating facts, sometimes even {\em contradict} the input article.

To make abstractive summarization models produce more factually consistent summaries, we need two critical components: an automatic evaluation metric for factual consistency and an effective training algorithm that maximizes factualness. Our main contributions lie in both areas. First, we propose an efficient automatic evaluation metric for factual consistency that is a simplification of the recently published QAGS protocol \cite{wang-etal-2020-asking}. Evaluating QAGS is computationally expensive and ill-suited for being part of the model training process. Our proposed protocol achieves a \textbf{55x} speedup while correlating closely with QAGS\footnote{See Sec. \ref{sec:appendix-speed} in the Appendix for details.}. Second, we propose a new contrastive learning method that uses factualness as a training objective. We demonstrate through experiments that our method improves the factual consistency of summarization models measured by both automatic metrics such as QAGS as well as human evaluation.

\section{An Efficient Metric for Factual Consistency} \label{sec:quals}
In order to improve factual consistency of summarization models, we must have a metric to quantify it. In addition, the metric needs to be computationally efficient so that we can incorporate it as part of the model training process. We first describe the QAGS protocol and then present our \textsc{Quals} protocol.
\begin{figure}%
	\centering
		\includegraphics[scale=0.18]{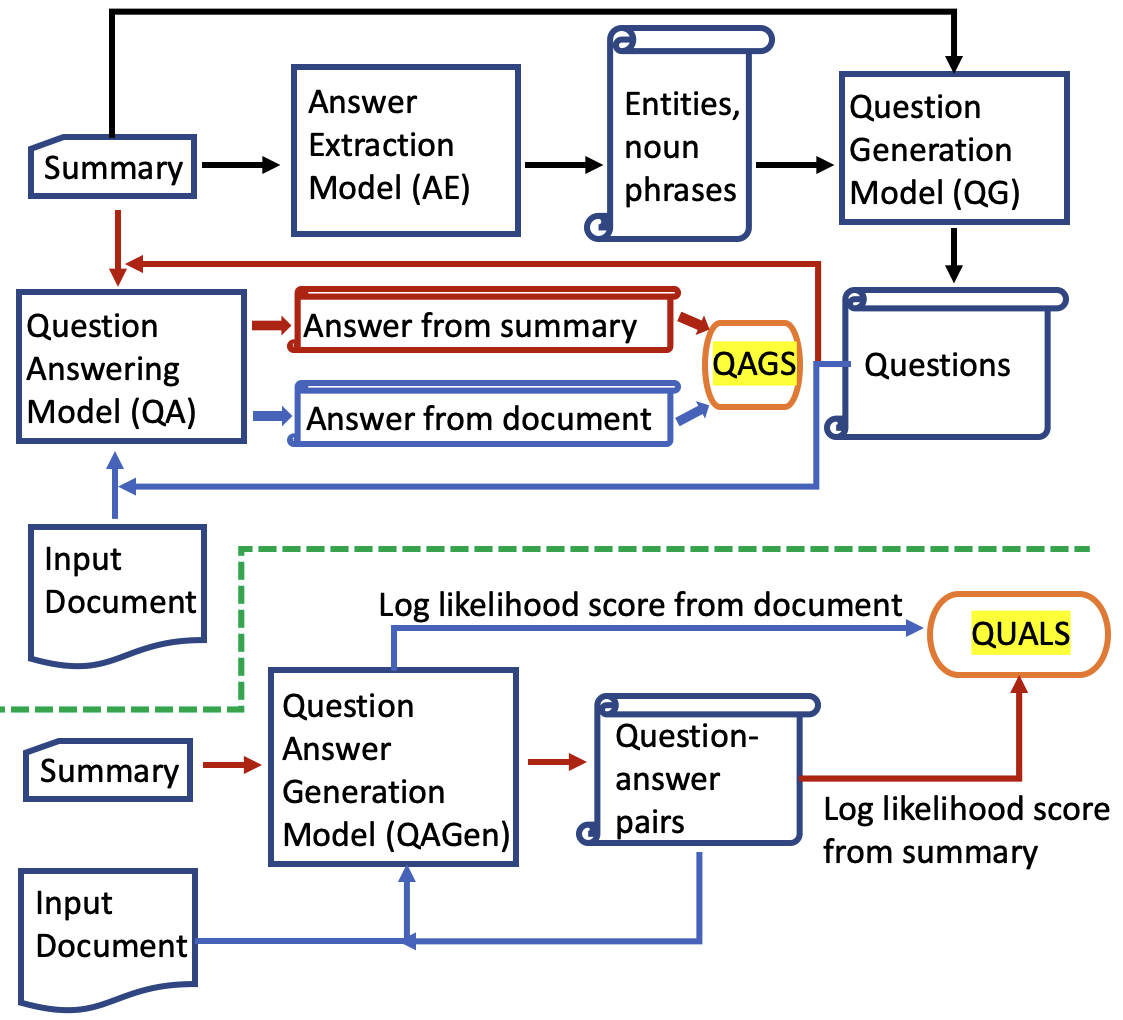}
	\caption{Comparison between \textsc{QAGS} (top) and \textsc{Quals} (bottom) protocols. \textsc{Quals} uses only one QAGen model instead of the AE, QG and QA models used in \textsc{QAGS}.}
	\label{fig:qags-compare}%
\end{figure}
\subsection{Background on \textsc{QAGS}}
Given a summary and an input document, QAGS \cite{wang-etal-2020-asking} scores the summary using a 4-steps pipeline: firstly, it extracts the named entities and noun phrases in the summary as candidate answers using an answer extraction (AE) model; secondly, a question generation (QG) model takes in the summary, concatenating with each candidate answer to generate a corresponding question; thirdly, a question answering (QA) model is used to answer each generated question in the context of the summary and the input document, separately; finally, the answers from the QA model based on the summary and the input document are compared to calculate F1 score in terms of their word level overlap as the QAGS score. Intuitively, for the same question, if the answer obtained from the input document matches that from the summary, it is an indication that the summary is factually consistent with the input document. We show the QAGS pipeline in the top part of Figure \ref{fig:qags-compare}. QAGS has the advantage of being interpretable and is shown to correlate well with human evaluation. However, using QAGS directly as a part of the training process presents several challenges. First, QAGS requires three separate models for AE, QG and QA. In addition to the summarization model being trained, these models 
consume a significant amount of machine memory. Second, performing these three steps separately takes a significant amount of time. For good coverage in QAGS, multiple answers are extracted for a given summary and multiple questions are generated for each answer. This means the QA model needs to perform inference on an exploding number of inputs even for one summary. Indeed, QAGS evaluation on a training set would take 584 days on a single GPU.\footnote{
See Sec. \ref{sec:appendix-speed} in the Appendix for details.} 
\subsection{\textsc{Quals} (ours)}
\begin{figure}%
	\centering
	\includegraphics[scale=0.15]{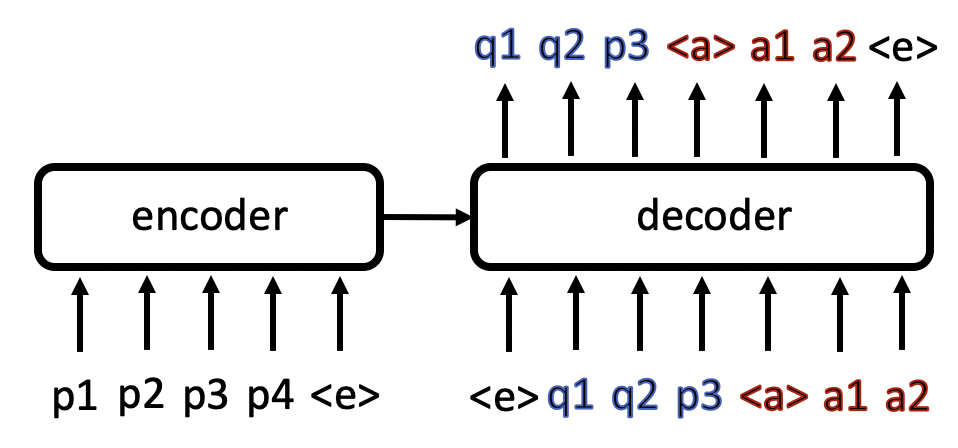}
	\caption{QAGen model: for an input text (p), it generates a question (q) followed by an answer (a).}
	\label{fig:qagen-diagram}%
\end{figure}
In order to enable the use of a QA driven metric to maximize factual correctness during the training of summarization models, we propose \textsc{Quals} (QUestion Answering with Language model score for Summarization),
which is illustrated in the bottom part of Figure \ref{fig:qags-compare}. 
QUALS is an efficient metric that employs a single neural language model (QAGen), as proposed in~\cite{shakeri-etal-2020-end}, to generate both the questions and answers from the summary. 
In particular, given a summary, QAGen outputs a question-answer (q-a) pair jointly, separated by a 
special token \verb|<a>| as shown in Figure \ref{fig:qagen-diagram}. Let $\text{LL}_{\text{summ}}(q,a)$ be the average log likelihood of generating the q-a pair from the given summary:
\begin{equation*}
\begin{footnotesize}
\begin{aligned}
 \text{LL}_{\text{summ}}(q,a) = & \frac{1}{N_q+N_a} \left( \sum_{i=1}^{N_q}\log p_{\text{QAGen}}(q^{i}|\text{summ}, q^{<i})\right. \\ 
 & \left. + \sum_{i=1}^{N_a}\log p_{\text{QAGen}}(a^{i} | \text{summ}, q, a^{<i}) \right), 
\end{aligned}
\end{footnotesize}
\end{equation*}
where $N_q$ and $N_a$ are the number of tokens for the question and answer, respectively. Note that we consider the log likelihood scores over both the question and answer tokens to account for factual consistency of both.
To obtain good coverage and diverse q-a pairs, we use diverse beam search \cite{DBLP:journals/corr/VijayakumarCSSL16} to generate 60 q-a pairs for a given summary with 60 diverse beam groups and a diverse beam strength of 0.5. We then filter out low-quality q-a pairs by keeping only those with answers found in the input summary. When multiple q-a pairs share the same answer, we only select the pair with the highest $\text{LL}_{\text{summ}}(q,a)$. 
Then given the input document, we simply evaluate the average log likelihood of the QAGen model producing the same q-a pairs, denoted as $\text{LL}_{\text{doc}}(q,a)$. 
Formally, given a summary and input document, \textsc{Quals} score is computed as follows:
\begin{equation*}
\begin{aligned}
 QUALS(\text{doc}, \text{summ}) = & \frac{1}{M} \sum^{M}_{i=1}{\left(\text{LL}_{\text{doc}}(q_i,a_i) \right.} \\
 & \left. - \text{LL}_{\text{summ}}(q_i,a_i)\right),
\end{aligned}
\end{equation*}
where $M$ is the number of q-a pairs selected on the summary.
There are two justifications for taking the difference between the log likelihood scores. 1. $\text{LL}_{\text{doc}}(q,a)$ alone only indicates the likelihood of the q-a pair given the document; subtracting $\text{LL}_{\text{summ}}(q,a)$ baselines it with the likelihood of generating the q-a pair given the summary. E.g. a low $\text{LL}_{\text{doc}}(q,a)$ does not necessarily imply factual inconsistency - it can be caused by the fact that the q-a pair itself is generated with low likelihood from the summary in the first place. 
2. Documents may vary in style, vocabulary and topic, which lead to variations in log likelihood scores unrelated to factual consistency; $\text{LL}_{\text{doc}}(q,a) - \text{LL}_{\text{summ}}(q,a)$ can help normalize these domain-related shifts since both the document and summary share the same basic style, vocabulary and topic. 

\section{Improving Factual Consistency Through Contrastive Learning} \label{sec:conseq}
Although QUALS can be computed more efficiently,
using it in the training process is not straightforward because one would need to backpropagate through generated sumaries and q-a pairs. We present our \conseq\ (CONtrastive SEQ2seq learning) algorithm that can effectively maximize such metrics in training. 

To fix notation, $x=x_1, \dots, x_m$ denotes a sequence of input tokens; $y=y_1, \dots, y_n$ denotes a sequence of target output tokens; $\hat{y} = \hat{y}_1, \dots, \hat{y}_{\hat{n}}$ denotes a sequence of generated tokens from a seq2seq model via sampling, i.e. $\hat{y} \sim p_{\theta}(\cdot | x)$, where $\theta$ is the parameter of the model.
Let $r(\hat{y}, x)$ be the evaluation (in our case the \textsc{Quals}) metric that we aim to maximize. 

\subsection{\conseq}
First, we train an initial seq2seq model with parameters $\theta^0$ using the original labeled training set $\{x^{(i)} , y^{(i)}\}$ via MLE. 
Second, we collect ground truth labeled training target sequences $y^{(i)}$ as well as the sampled sequence $\hat{y}^{(i)}$ to form a set of candidate sequences $\mathcal{S} = \{y^{(i)}, \hat{y}^{(i)} \}$. 
Third, we construct $\mathcal{S}^+$ and $\mathcal{S}^-$ from $\mathcal{S}$ based on the evaluation scores $r$ and minimize the following loss function from the initial parameters $\theta^0$:
\begin{align}\label{eq:contrastive}
\mathcal{L}_{\mathtt{contrast}} = & \underbrace{-\mathbb{E}_{x,s \in \mathcal{S}^+} \log p_{\theta}(s | x)}_{\mathcal{L}_{\mathtt{contrast}}^+} 
\\ \notag & \underbrace{-\mathbb{E}_{x,s \in \mathcal{S}^-} \log \left(1-p_{\theta}(s | x)\right)}_{\mathcal{L}_{\mathtt{contrast}}^-}.
\end{align}
Intuitively, $\mathcal{S}^+$ consists of highly rewarded sequences (factually consistent summaries) and minimizing $\mathcal{L}_{\mathtt{contrast}}^+$ forces the model to generate high score sequences; likewise, 
$\mathcal{S}^-$ consists of poorly rewarded sequences (factually inconsistent summaries) and minimizing $\mathcal{L}_{\text{contrast}}^-$ forces the model to move away from low score sequences.
We present the full method in Algorithm \ref{algo:contrastive}. 
\begin{algorithm} 
	\SetAlgoLined
	\KwIn{Initial seq2seq (summarization) model weights $\theta^0$ via MLE, input and target sequences $\{x^{(i)}, y^{(i)}\}$, evaluation metric $r$.}
	Initialize $k=0$\;
	\While{not converged}{
		Sample candidate sequences $\{ \hat{y}^{(i)} \}$ for input sequences $\{x^{(i)}\}$ and $\theta^k$\;
		Construct $\mathcal{S}^+$ and $\mathcal{S}^-$ as described in Sec. \ref{sec:conseq}\;
		Minimize the contrastive loss in Eq. \ref{eq:contrastive} to obtain $\theta^{k+1}$\;
		$k = k + 1$\;
	}
	\caption{\conseq}\label{algo:contrastive}
\end{algorithm}
\paragraph{Comparison with REINFORCE: }
The typical approach to directly optimize a non-differentiable evaluation score during training is the REINFORCE algorithm \cite{williams1992simple}. 
REINFORCE samples a sequence $\hat{y}$ at each iteration and updates the model with the gradient
\begin{equation}\label{eq:reinforce}
\left( r(\hat{y}, x) - r(b, x)\right) \bigtriangledown_{\theta} \log p_{\theta}(\hat{y} | x),
\end{equation}
where $b$ is a baseline sequence, conditionally independent of $\hat{y}$ given $\theta, x$.
To see the connection with \conseq, suppose the reward $r$ is either 0 or 1. If $r(\hat{y}, x) = 1$ and $r(b, x) = 0$, the sampled sequence $\hat{y}$ is strongly rewarded compared to baseline and Eq. \ref{eq:reinforce} reduces to $\bigtriangledown_{\theta} \log p_{\theta}(\hat{y} | x)$. On the other hand, if $r(\hat{y}, x) = 0$ and $r(b, x) = 1$, the sampled sequence is strongly discouraged and Eq. \ref{eq:reinforce} reduces to $-\bigtriangledown_{\theta} \log p_{\theta}(\hat{y} | x)$, which pushes the model away from generating $\hat{y}$. This pull-and-push  effect is analogous to the ${\mathcal{L}_{\mathtt{contrast}}^+}$ and ${\mathcal{L}_{\mathtt{contrast}}^-}$ terms in the loss Eq. \ref{eq:contrastive} in \conseq.
Note that the gradient updates of REINFORCE are entirely based on the sampled sequences. In contrast, \conseq\ takes advantage of the ground truth targets in addition to the sampled ones, which help avoid the instability of REINFORCE. 
Indeed, we implemented the REINFORCE algorithm with the BART-large model fine-tuned under MLE objective as initialization; we found that after a few hundred updates the summaries sampled from the model become unintelligible and our reward function fails to compute the scores (no meaningful q-a pairs can be generated based on the summaries). 

\subsection{\conseq\ + \textsc{Quals} for Imposing Factual Consistency}
We use \textsc{Quals} to select high quality positive and negative examples for \conseq\ with the goal of training seq2seq summarization models that are more factual.
In order to create $\mathcal{S}^+$ and $\mathcal{S}^-$ we first evaluate \textsc{Quals} for all the ground truth summaries of the training set and select $p$\% of those with the highest \textsc{Quals} scores to form $\hat{\mathcal{S}}^+$.\footnote{We found it necessary to select the top $p$\%  of the ground truth summaries to form $\hat{\mathcal{S}}^+$ because not all ground truth summaries are factually consistent to the input documents, due to the imperfect data collection process. This is especially true for the XSUM dataset as we discuss in the next section.}
To generate the negative samples, we use the $\text{topK}$ sampling ($k=50$) during decoding to generate 6 summaries for each input document in the training set; we then select the one with the lowest \textsc{Quals} score out of the 6 summaries for each input document; next, we rank the selected summaries and choose $p$\% of those with the lowest \textsc{Quals} scores to form $\hat{\mathcal{S}}^-$. Note that the summaries in $\hat{\mathcal{S}}^+$ and $\hat{\mathcal{S}}^-$ may correspond to different input documents. The last step is to take the intersection of the examples between $\hat{\mathcal{S}}^+$ and $\hat{\mathcal{S}}^-$ to form $\mathcal{S}^+$ and $\mathcal{S}^-$, respectively. For example, we select a summary $s$ from $\hat{\mathcal{S}}^+$ to be included in $\mathcal{S}^+$ if and only if there exists a summary $s'$ in $\hat{\mathcal{S}}^-$ such that $s$ and $s'$ correspond to the same input document. 
As a result of the above process, the contrastive loss in Eq. \ref{eq:contrastive} can thus push the model from the inconsistent summary towards the consistent
one for the same input document. Next, we describe two variants of the \conseq\ algorithm.

\paragraph{Weighted loss:} We can weight the losses in Eq. \ref{eq:contrastive} using \textsc{Quals} scores and minimize the following loss, assuming normalization of $0\leq r \leq 1$:
\begin{align*}
& \mathcal{L}_{\text{contrast}} =  -\mathbb{E}_{x,s \in \mathcal{S}^+} r(s, x) \log p_{\theta}(s | x) \\ & -\mathbb{E}_{x,s \in \mathcal{S}^-} \left(1-r(s, x)\right) \log \left(1-p_{\theta}(s | x)\right),
\end{align*}
where $r(s, x)$ is the \textsc{Quals} score for summary $s$ and input document $x$.

\paragraph{Online learning:} We refer to Algorithm \ref{algo:contrastive} as the offline training setting in the sense that in each iteration, $\mathcal{S}^+$ and $\mathcal{S}^-$ are constructed by pooling together all available input documents and their candidate summaries to train the model.
It is also possible to perform training in an online fashion. Specifically, we can take in a batch of input sequences in each iteration, construct $\mathcal{S}^+$ and $\mathcal{S}^-$ based only on the examples in the batch, and take a gradient step with respect to Eq. \ref{eq:contrastive}. Compared to the offline setting, the model parameters are updated much more frequently and the candidate sequences are always generated from the latest model parameters. On the other hand, the construction of $\mathcal{S}^+$ and $\mathcal{S}^-$ are restricted to the examples within the batch, resulting in potentially less representative samples compared to the offline setting. 

\section{Experiments}
\subsection{Experimental Setup}
\paragraph{Datasets:}
We perform our summarization experiments on two widely used news datasets: XSUM \cite{narayan-etal-2018-dont} and CNN/DailyMail \cite{nallapati-etal-2016-abstractive}.
The XSUM dataset consists of short, one-sentence summaries of the BBC news articles. The dataset is constructed by taking the first sentence of an article as the summary and the rest of the article as input document. As a result, the summaries are highly abstractive. At the same time, there are many examples where a summary contains information (e.g. the first name of a person) that is not mentioned in the input document. This introduces an undesirable bias in the training data to encourage the model to hallucinate. 
The CNNDM dataset contains multi-sentence (4 sentences on average) summaries of news articles from the CNN and DailyMail. The summaries are curated by human annotators in terms of highlights of the article. Compared to XSUM, the summaries in CNNDM are much more extractive - each summary sentence usually corresponds to an existing sentence in the input document. 

\paragraph{Evaluation metrics:} We use the ROUGE \cite{lin2004rouge} to measure general summarizaiton quality. 
For factual consistency, we use the QAGS protocol (see Appendix for more details) as well as the FactCC model \cite{kryciski2019evaluating} downloaded directly from the official website.\footnote{\url{https://github.com/salesforce/factCC}}  
In contrast to QAGS,
FactCC is a BERT-based classification model that makes a binary prediction if the given claim sentence is factually consistent or not with the given input document. 

\paragraph{Implementation details:} We use the \texttt{Fairseq} \cite{ott2019fairseq} implementation of BART-large \cite{lewis2019bart} for the summarization model as it is shown to achieve the state-of-the-art ROUGE scores for this task. We fine-tune the BART-large model with the standard learning rate of $3\times 10^{-5}$ on XSUM and CNNDM respectively to establish the MLE baselines. We then initialize \conseq\ with the MLE baseline models. In \conseq\ we use a learning rate of $3\times 10^{-6}$. For evaluation, we generate summaries using beam search with beam sizes of 4 and 6 for CNNDM and XSUM, respectively. The generated summaries are limited to 55-140 and 10-60 tokens in lengths for CNNDM and XSUM, respectively. Our QAGen model in \textsc{Quals} is also a BART-large model fine-tuned on the \textsc{SQuad} \cite{rajpurkar2016squad} and NewsQA \cite{trischler2017newsqa} datasets. To construct the $\mathcal{S}^+$ and $\mathcal{S}^-$, we found that selecting the $p=30$\% and $50$\% leads to the best result on the validation set of XSUM and CNNDM, respectively, among the choices of $p=25, 30, 50, 75, 90$.

\subsection{\textsc{Quals} Approximates \textsc{QAGS}}
We first verify that our proposed \textsc{Quals} metric correlates well with QAGS. We evaluate both \textsc{Quals} and \textsc{QAGS} on the same set of summaries generated by the MLE baseline model on the test set of documents in XSUM and CNNDM, respectively. The examples are grouped into bins based on the percentiles of the \textsc{Quals} scores. We then plot the average \textsc{QAGS} score of the examples within each bin. As shown in Figure \ref{fig:qags-approx-bar} 
(a more fine-grained plot is shown in Figure \ref{fig:qags-approx} of the Appendix), \textsc{Quals} correlates very well with \textsc{QAGS} in both datasets. 
Since our method only relies on ranking QUALS scores in contrastive learning,
monotonicity of QUALS with respect to QAGS is sufficient.

\begin{figure}%
	\centering
    \includegraphics[width=0.45\textwidth,height=0.155\textheight]{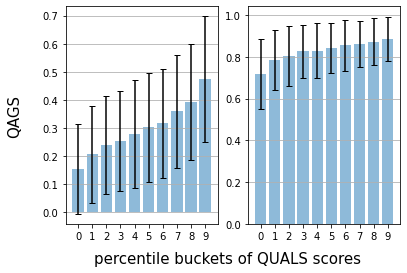}
	\caption{Correlation between \textsc{Quals} and \textsc{QAGS} on XSUM (left) and CNNDM (right). The average \textsc{QAGS} tend to increase with the increase in \textsc{Quals}.}
	\label{fig:qags-approx-bar}%
\end{figure}

\subsection{Results}
We compare our proposed method \textsc{Quals}-\conseq\ to the state-of-the-art abstractive summarization model (BART-large MLE). 
In an ablation study, we check the effect of changing the \textsc{Quals} metric 
as well as the effect of changing the \conseq\ algorithm. We summarize the results in Table \ref{tab:xsum} and Table \ref{tab:cnndm}. 
We observe that our proposed method \textsc{Quals}-\conseq\ (\textbf{Q-C}) achieves more than 4 points improvement in QAGS over the MLE baseline in XSUM and about 2 points improvement in CNNDM, where we also achieve a slightly better ROUGE over MLE. 
Improving ROUGE is not the goal of our paper; what we show is that we can significantly improve factual consistency of summaries without degrading ROUGE, as is common practice~\cite{kedzie-mckeown-2019-good}. Next, we describe the various ablation settings.

{\bf 1)} In \textbf{R-C} (ROUGE-\conseq), we simply use the sum of ROUGE-1,2,L scores to evaluate the generated summaries against the ground truth summaries as the metric in constructing $\mathcal{S}^+$ and $\mathcal{S}^-$. In both Table \ref{tab:xsum} and \ref{tab:cnndm} it results in poorer QAGS than the MLE baseline. This confirms the necessity of having an effective metric for factual consistency. Note that \textbf{R-C} even results in poorer ROUGE scores. We believe this is caused by the fact that ROUGE is already highly optimized by the MLE model and it is used as initialization for \textbf{R-C}; the ``hard'' examples where the MLE model couldn't produce good ROUGE scores may be inherently problematic (e.g. hallucination in the ground truth summary); focusing on these examples by \textbf{R-C} can therefore make the model weaker on other examples.

{\bf 2)} In  \textbf{Q-F1-C} (\textsc{Quals}-F1-\conseq), we make a modification to \textsc{Quals}. Instead of measuring the factual consistency in terms of the log likelihood scores, we measure the F1 between generated answers from the summary and the input document in the QAGen model. In particular, given a summary as input, the QAGen model generates a q-a pair $q,a$. We then use the corresponding document as input to the QAGen model and force the decoder to generate the question tokens $q$ and allow the QAGen to generate the answer tokens $a'$.  We then compute the F1 overlap score between $a$ and $a'$. This would be closer to the QAGS setting where explicit answers are generated and compared. We observe that in Table \ref{tab:cnndm}, \textbf{Q-F1-C} achieves a slightly higher QAGS than \textbf{Q-C}. But overall \textbf{Q-F1-C} performs worse than \textbf{Q-C}. We believe this is due to the fact the log likelihood scores are softer than F1 and can potentially account for answers that are semantically similar. 

{\bf 3)} In \textbf{Q-C-W} (\textsc{Quals}-\conseq-Weighted), we use the weighted version of \conseq\ as described in Sec. \ref{sec:conseq}. Since the \textsc{Quals} score is a difference between log likelihood scores, it can have negative values. We evaluate the \textsc{Quals} on the training examples to obtain an interval of its values and linearly normalize the \textsc{Quals} as weights in the loss function. We observe that it improves the factual consistency over the MLE baseline but not as much as \textbf{Q-C}.

{\bf 4)} In \textbf{Q-C-O} (\textsc{Quals}-\conseq-Online), we use the online version of \conseq\ as described in Sec. \ref{sec:conseq}. We sample about 6 examples in a mini-batch and select 2 of them for $\mathcal{S}^+$ and $\mathcal{S}^-$ per GPU with a total of 40 GPUs. We observe that it tends to achieve higher ROUGE scores but lower factual consistency scores compared to \textbf{Q-C}.

{\bf 5)} In \textbf{Q-P} (\textsc{Quals}-Positive), we only use the positive summaries ($\mathcal{S}^+$) and the positive loss $\mathcal{L}_{\mathtt{contrast}}^+$ in Eq. \ref{eq:contrastive} for training. We observe that it achieves lower factual consistency scores compared to \textbf{Q-C} and this shows that the negative loss in \conseq\ is useful to boost factual consistency.

\begin{table}[]
	\centering
	\resizebox{0.49\textwidth}{!}{%
		\begin{tabular}{@{}c|c|c|c|c|c|c@{}}
			\toprule
			& \textsc{Quals} & QAGS   & FactCC & ROUGE 1 & ROUGE 2 & ROUGE L \\ \midrule
			MLE       & -1.0393 & 29.77 & 23.64 & 45.18 & 22.19 & 36.97 \\ \midrule
			R-C & -1.0907  & 28.47 & 23.76 & 44.59 & 21.88 & 36.58 \\ \midrule
			Q-F1-C & -0.9866 & 32.76 & 22.75 & 44.54 & 21.63 & 36.37 \\ \midrule
			Q-C-W & -0.9856      & 31.39       & 21.68    & 44.42   & 21.17       & 35.94       \\ \midrule
			Q-C-O & -0.9747     &  32.26       & {\bf 23.68}    & {\bf 45.22}   & {\bf 22.19}       & {\bf 37.00}       \\ \midrule
			Q-P & -0.9739      & 31.92       & 22.68    & 45.02   & 22.00       & 36.83       \\ \midrule			
			Q-C    & {\bf -0.9061} & {\bf 34.36} & 22.42 & 44.67 & 21.66  & 36.47 \\ \bottomrule
		\end{tabular}%
	}
	\caption{Test set results on XSUM. \textsc{Quals}-\conseq\ (Q-C) achieves over 4 points higher QAGS than the BART-large MLE baseline.}
	\label{tab:xsum}
\end{table}

\begin{table}[]
	\centering
	\resizebox{0.49\textwidth}{!}{%
		\begin{tabular}{@{}c|c|c|c|c|c|c@{}}
			\toprule
			& \textsc{Quals} & QAGS    & FactCC & ROUGE 1 & ROUGE 2 & ROUGE L \\ \midrule
			MLE       & 0.0169 & 82.84 & 68.33 & 44.24 & 21.35 & 41.18 \\ \midrule
			R-C & -0.0701      & 79.28       & 63.50    & 41.17   & 18.41       & 37.94       \\ \midrule
			Q-F1-C &  0.0779      &   {\bf 84.97}     & 70.54    & 44.23   & 21.24       & 40.96       \\ \midrule
			Q-C-W & 0.0720      & 83.54       & 70.03    & 44.04   & 20.88       & 40.74       \\ \midrule		Q-C-O & 0.0292      & 83.08       & 68.16    & {\bf 44.70}   & 21.58       & {\bf 41.53}       \\ \midrule
			Q-P & 0.0437     & 83.84       & 69.40    & 44.62   & {\bf 21.65}       & 41.47       \\ \midrule			
			Q-C    & {\bf 0.0857}   & 84.75 & {\bf 72.83} & 44.40 & 21.37 & 41.17 \\ \bottomrule
		\end{tabular}%
	}
	\caption{Test set results on CNNDM. \textsc{Quals}-\conseq\ (Q-C) achieves about 2 points higher QAGS than the BART-large MLE baseline.}
	\label{tab:cnndm}
\end{table}

\paragraph{FactCC results:}
 As shown in Table \ref{tab:cnndm} for CNNDM, \textbf{Q-C} achieves over 4 points improvements in FactCC score over the MLE baseline. However, in Table \ref{tab:xsum} for XSUM, \textbf{Q-C} has about 1 point lower FactCC score than the MLE baseline. We investigated this issue and found that the ground truth summaries of the XSUM test set have a FactCC score of just 21.0, which means that only 21\% of the ground truth summaries in XSUM are judged as factual according to FactCC. This suggests that the FactCC model is not well suited for making predictions on highly abstractive summaries. This is not surprising as the authors of FactCC mentioned in Sec. 3.1 \cite{kryciski2019evaluating} that FactCC is built on the premise that ``..the level of abstraction of generated summaries is low and models mostly paraphrase single sentences and short spans from the source''. Unfortunately for XSUM, this premise does not hold.
\paragraph{Comparison with Other Methods:}
There are 2 other methods in the literature \cite{AAAI1816121, zhu2020boosting} for improving factual consistency of summarization models. Both rely on information extraction (OpenIE) to extract relations and incorporate the relation representations into seq2seq models. The authors in \cite{zhu2020boosting} proposed a Fact-Aware Summarizer (FASum) and a Fact Corrector model. In Table 4 of their paper, the FASum achieves significantly lower ROUGE scores (30.28/10.03/23.76 and 40.53/17.84/37.4 for ROUGE-1/2/L on XSUM and CNNDM respectively). This indicates a significant gap in the summary quality.
Even their best result, which is using Fact Corrector on UniLM \cite{DBLP:conf/nips/00040WWLWGZH19}, achieves lower ROUGE scores than BART-large MLE. 
Although the authors in \cite{zhu2020boosting} used FactCC as an evaluation metric, they did not use the official method to train FactCC; they used the ground truth summaries rather than sampled sentences from the input documents as positive examples. As a result, we are not able to compare the FactCC numbers reported in \cite{zhu2020boosting}. Nevertheless, we can 
observe that there is little or no improvements for Fact Corrector on UniLM according to FactCC. We believe that this is because the recent large transformer-based, pre-trained seq2seq models such as UniLM and BART have significantly improved the summarization 
quality and it is much more challenging to improve even the factual consistency of these state-of-the-art models. In comparison, our results reported in Table \ref{tab:xsum} and Table \ref{tab:cnndm} represent  significant improvements. The authors in \cite{AAAI1816121} only experimented on the Gigaword corpus \cite{rush-etal-2015-neural} and did not release their code so we were unable to compare to their method. However, given the recent progress in transformer-based  seq2seq models, it is likely that our BART-large MLE baseline outperforms their RNN-based models. Again, we believe that it is much easier to improve factual consistency of a weak seq2seq model than that of a strong model (such as UniLM or BART-large) as shown in \cite{zhu2020boosting}.

\paragraph{Human evaluation:}
We use Amazon SageMaker Ground Truth\footnote{\url{https://aws.amazon.com/sagemaker/groundtruth/}} to conduct human evaluation. We sample 100 examples from the test set of XSUM and CNNDM, respectively. In each task, we present an input document, together with the generated summaries from the BART-large MLE and \textsc{Quals}-\conseq\ models. We ask the annotators to select which of the two summaries they prefer along 3 dimensions: factual consistency, informativeness and grammatical correctness. For each of these dimensions they can also choose ``Equal'' if they feel that both summaries are of similar quality. 
Our annotators consist of 10 data associates who are native English speakers whose background includes training in linguistic annotation. Each task is performed by 3 different annotators and we take the majority vote. 
We provide the detailed 
setup and instructions in the Appendix. 

The result of human evaluation is reported in Table \ref{tab:human-eval-results}, showing the percentage of examples along these three dimensions.
\begin{table}
	\centering
	\small
	\resizebox{.5\textwidth}{!}{
		\begin{tabular}{lccccccccc}
			\toprule
			\multirow{2}{*}{Metrics} & \multicolumn{3}{c}{Factual} &  \multicolumn{3}{c}{Informative} & \multicolumn{3}{c}{Grammatical}  \\
			\cmidrule(lr){2-4} \cmidrule(lr){5-7} \cmidrule(lr){8-10} 
			& {better} & {worse} & {equal} & {better} & {worse} & {equal} & {better} & {worse} & {equal} \\
			\midrule
			XSUM & 18 & 9 & 73 & 22 & 9 & 69 & 4 & 2 & 94  \\
			CNNDM & 18 & 7 & 75 & 42 & 22 & 36 & 5 & 6 & 89 \\ 
			\bottomrule
		\end{tabular}
	}
	\caption{Human evaluation results on summaries generated by \textsc{Quals}-\conseq\ in comparison to the BART-large MLE baseline for 100 randomly selected examples from the test sets of XSUM and CNNDM. 
	}\label{tab:human-eval-results}
\end{table}
In both datasets, we observe that \textsc{Quals}-\conseq\ clearly improves the factual consistency of the generated summaries compared to the BART-large MLE baseline. We notice that the improvement in informativeness is even greater. 

Fleiss's Kappa~\cite{fleiss1971mns} shows fair agreement for  factual consistency, informativeness and grammatical correctness choices (0.136/0.270/0.043 for XSUM and 0.237/0.202/0.206 for CNNDM). We note, however, that most disagreements occur when one annotator rates two summaries as equal and another rates one of the two as either better or worse. To measure this, we computed Fleiss's Kappa again, counting equal and either better or worse as equivalent (and better and worse as not equivalent). Here, our agreement is almost perfect (0.837/0.839/0.975 for XSUM and 0.945/0.816/0.967 for CNNDM). We thus see that annotators rarely directly contradict each other on rating one summary above or below another, but often have a hard time deciding when the two summaries are equal. 

\subsection{Qualitative Analysis}
We analyzed the human evaluation results and found several types of improvements/errors produced by \textsc{Quals}-\conseq. Our model is able to rectify factual errors found in MLE such as 1) entity hallucination and errors (Example 1 and 2 in Table \ref{tab:qualitative_analysis}) and 2) relations and co-reference (see Table \ref{tab:qualitative_results} and Example 3 in Table \ref{tab:qualitative_analysis}). QUALS-CONSEQ also made mistakes in cases where it was not sensitive to certain modifier phrases (extra ``more than'' in Example 2 
in Table \ref{tab:qualitative_analysis}). More examples of generated summaries and q-a pairs are in the Appendix.
\begin{table}[]
	\centering
	\resizebox{0.5\textwidth}{!}{%
		\scriptsize
		\begin{tabular}{p{0.1\linewidth}p{0.8\linewidth}}
			\toprule
			Input 1: & Keates made over 150 league appearances for Wrexham and captained the club to an FA Trophy win in 2013. ... His first game in charge as permanent manager will be ... \\
			MLE: &  Wrexham have appointed Dean Keates as their new manager on a \textcolor{red}{two-year contract}. \\
			\conseq: & Wrexham have appointed former captain Dean Keates as their new manager.
			\\  \cmidrule(l){1-2}
			Input 2: & Passwords were found on public websites such as Pastebin, where hackers often dump data. ... It found 705 emails and passwords originating from government agencies. ... \\
			MLE: & \textcolor{red}{More than} \textcolor{red}{700,000} government emails and passwords have been leaked online ... \\
			\conseq: & \textcolor{red}{More than} 705 emails and passwords belonging to US government agencies have been found on the open web ... \\ \cmidrule(l){1-2}
			Input 3: & The girlfriend of a British student killed in the Alps plane tragedy ... revealed she did not blame the co-pilot who crashed the jet. Paul Bramley, 28, died when Andreas Lubitz locked the Germanwings flight's captain out of the cockpit before flying the plane into a mountainside... \\
			MLE: & Paul Bramley, 28, died when Andreas Lubitz locked \textcolor{red}{him} out of cockpit. ...
			\\ 
			\conseq: & Paul Bramley, 28, died when Andreas Lubitz locked captain out of cockpit. ...	
			\\ 
			\bottomrule
		\end{tabular}%
	}
	\caption{Qualitative analysis. Example 1: ``two-year contract'' was never mentioned in the input document. Example 2: ``700,000'' was wrong in MLE output; ``more than'' was inaccurate. Example 3: Andreas Lubitz locked the captain out of cockpit, not Paul Bramley. 
	}
	\label{tab:qualitative_analysis}
\end{table}

\paragraph{Illustration of \textsc{Quals}:}
We take an example to illustrate how \textsc{Quals} captures the factual inconsistency of summaries. The BART-large MLE model generate a summary: \emph{The AirAsia flight \textcolor{red}{4U 9525} crash was the latest in a series of tragedies that have hit the aviation industry.}
The input document described the AirAsia crash but did not mention the flight number. In fact, ``4U 9525'' is the Germanwings flight that crashed in the French Alps. The model hallucinated the flight number because it appeared in several training examples that cover the Germanwings crash. Given the above summary, our QAGen model generates the following q-a pairs: \emph{Q1: What was the name of the flight that crashed? A1: 4U 9525. Q2: Which airlines flight crashed? A2: AirAsia}. In Figure \ref{fig:log-likelihood-histogram} in the Appendix we show the negative log likelihood per subword token on these q-a pairs conditioned on the summary (blue) and input document (orange). The answer to the first question is very likely according to the summary while extremely unlikely according to the input document, indicating factual inconsistency. On the other hand, ``AirAsia'' is factually consistent and the second q-a pair is likely according to the input document. The \textsc{Quals} score for the two q-a pairs are $-2.615$ and $-0.054$, respectively.

\section{Related work}
Several authors have pointed out the problem of factual inconsistency in abstractive summarization models \cite{kryscinski-etal-2019-neural, AAAI1816121, durmus2020feqa}. 
Besides QAGS \cite{wang-etal-2020-asking} and FactCC \cite{kryciski2019evaluating}, another possible approach to quantify factual consistency is to rely on Open Information Extraction (OpenIE) and dependency parsing tools to identify and match the relations in an input document and its summary \cite{AAAI1816121, zhu2020boosting}. However, the underlying OpenIE tools are often not accurate enough to be used for this purpose. 

Our proposed \conseq\ algorithm is related to the unlikelihood training \cite{welleck2019neural, DBLP:journals/corr/abs-1911-03860} as both have positive and negative loss terms. The key difference is that in unlikelihood training, the negative loss serves as a regularization term, weighted by a hyperparameter $\alpha$, in addition to the regular MLE training. In contrast, our \conseq\ is motivated from the REINFORCE algorithm and treats the positive and negative terms equally. Furthermore, while the unlikelihood training uses all the ground truth sequences equally in the regular MLE (positive) loss term, we construct the positive and negative sets by incorporating the reward function (e.g. \textsc{Quals}) as discussed in Sec. \ref{sec:conseq}. 

In another related work, factual consistency metrics at the entity level have been proposed \cite{nan-etal-2021-entity}. The authors also investigated several techniques such as data cleaning, multi-task learning and entity-augmented decoding to improve entity level factual consistency scores of abstractive summarization models. In contrast, the \textsc{Quals} metric that we propose is more general, not limited to entities. 
Another recent work tackles the hallucination problem in abstractive text summarization via post processing on the generated summary \cite{chen2021improving}. Specifically, entities of the generated summaries are swapped with other named entities of the same type found in the original document to form a set of candidate summaries. The final summary is determined by a ranking model trained to prefer the factually consistent summaries.

\section{Conclusion}
In this paper we proposed to improve the factual consistency of abstractive summarization models. We first proposed an efficient evaluation protocol called \textsc{Quals} to measure factual consistency. We then proposed a contrastive learning algorithm for seq2seq models called \conseq\ to maximize \textsc{Quals} during training. We demonstrated that our proposed method significantly improves the factual consistency of the current state-of-the-art summarization model measured by  automatic metrics as well as side-by-side human evaluation. In addition to improving factual consistency of summarization models, we believe that the \conseq\ algorithm can have a wider impact on training seq2seq models in general to incorporate non-differentiable evaluation metrics into model training.

\bibliography{contrastive_seq2seq}
\bibliographystyle{acl_natbib}

\clearpage

\appendix

\section{Appendices}
\label{sec:appendix}

\subsection{Our implementation of QAGS}
For answer extraction, we follow the original authors to use Spacy to extract named entities and noun phrases. We filter out the stop words such as ``who'', ``it'' as we find them uninformative in question generation and keep 10 answer candidates. We then use the BART-large model trained on NewsQA \cite{trischler2017newsqa} dataset for question generation. For each answer candidate we use beam size of 10 and select the top 3 questions per answer. So we would have 30 questions per summary. For the question answering model, we use the ALBERT-xxlarge \cite{lan2019albert} model trained on \textsc{SQuad2.0} \cite{rajpurkar2018know} as it achieves even better accuracy than BERT-large.

\subsection{Speed Estimate for \textsc{Quals} and QAGS}\label{sec:appendix-speed}
Evaluating the QAGS using the ALBERT-xxlarge as the QA model on the test set (11490 examples) of CNNDM would take about 93.6 hours on a single NVIDIA V100 Tensor Core GPU. \textsc{Quals} only takes about 1.7  hours on the same GPU, offering more than 55x speedup. If we were to use QAGS during training (287112 examples in the training set), we would need to evaluate it for each training example 6 times (6 sampled candidate summaries). It would take a staggering 14033 hours, or 584 days on a single GPU. On the other hand, we are able to compute \textsc{Quals} on a machine with 4 GPUs for the training set in 66 hours. We believe the efficiency of \textsc{Quals} is critical in enabling the optimization of factual consistency.

\subsection{Additional Experimental Details of \conseq}
We experimented with other ways of constructing $\mathcal{S}^+$ and $\mathcal{S}^-$. We tried using $\hat{\mathcal{S}}^+$ and $\hat{\mathcal{S}}^-$ directly in \conseq\ (without taking intersection of examples) and found that the overall quality of the summaries are worse than taking the intersection. After taking the intersection, the sizes $|\mathcal{S}^+| = |\mathcal{S}^-|=6492,48809$, respectively for XSUM and CNNDM. Note the original training sets have 203540 and 287112 examples for XSUM and CNNDM, respectively. Furthermore, we found that using the positive and negative summaries corresponding to the same document in a minibatch leads to better results. We also found in our experiments that the best results are obtained with only one outer iteration of Algorithm \ref{algo:contrastive}. 

\subsection{\textsc{Quals} Approximates \textsc{QAGS}}
\begin{figure}%
	\centering
    \includegraphics[width=0.45\textwidth,height=0.155\textheight]{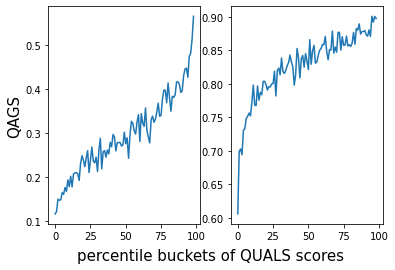}
	\caption{Correlation between \textsc{Quals} and \textsc{QAGS} on XSUM (left) and CNNDM (right). The average \textsc{QAGS} tend to increase with the increase in \textsc{Quals}. The standard deviation of the QAGS for each bin is about 0.187 for XSUM and 0.127 for CNNDM.}
	\label{fig:qags-approx}%
\end{figure}

As shown in Figure \ref{fig:qags-approx-bar} in the main paper, \textsc{Quals} correlates very well with \textsc{QAGS} in both datasets with 10 percentile bins. We also show the correlation plot with 100 bins in Figure \ref{fig:qags-approx}, with the same monotonicity trend.

\subsection{Illustration of \textsc{Quals}:}
\begin{figure}%
	\centering
	\subfigure{%
		\includegraphics[width=0.5\textwidth,height=0.159\textheight]{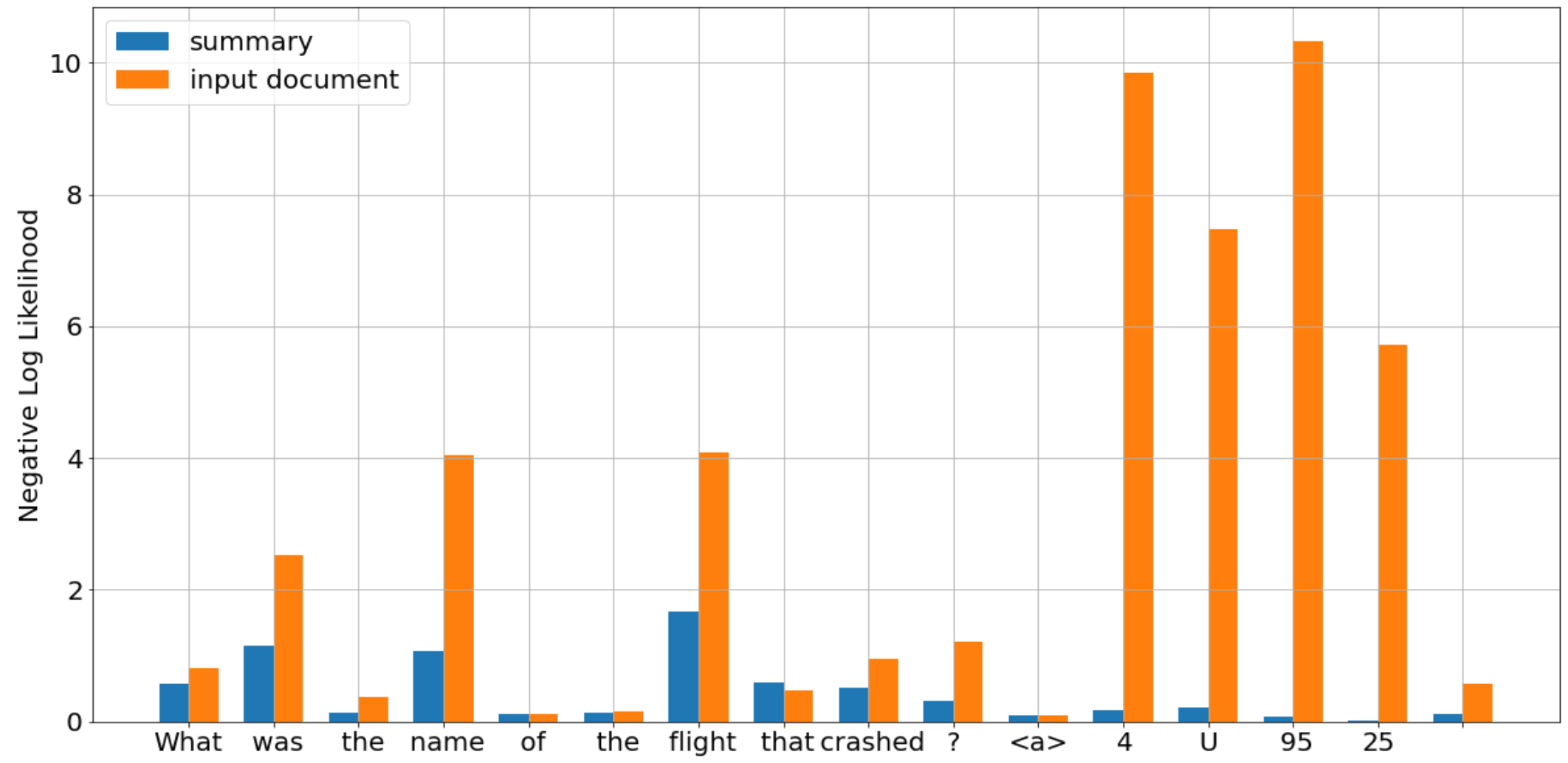}}\\
	\subfigure{%
		\includegraphics[width=0.5\textwidth,height=0.159\textheight]{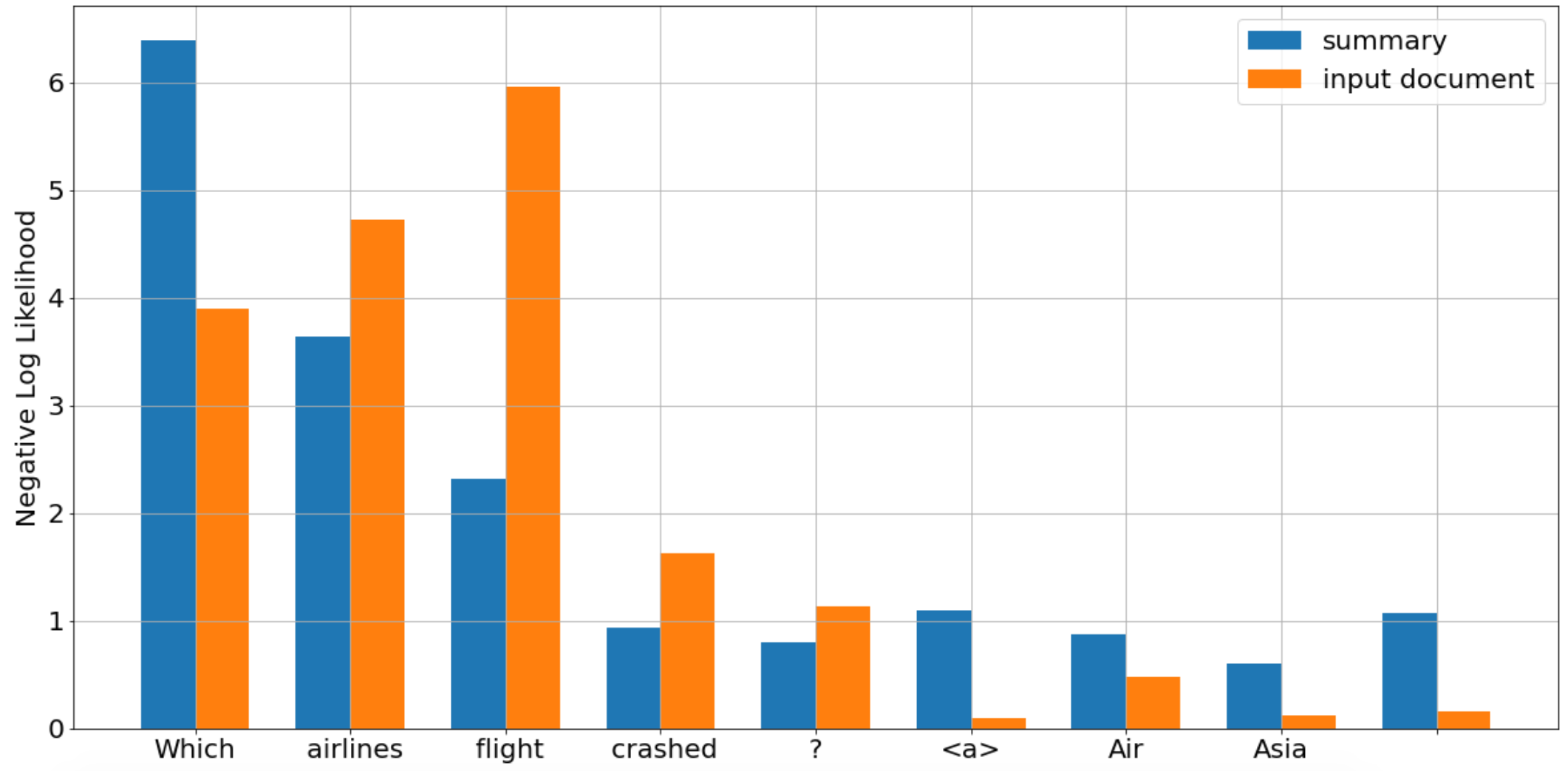}}
	\caption{Negative log likelihood per subword token on two q-a pairs from the QAGen model according to the summary(blue) and input document (orange). Higher means unlikely. The first q-a pair (top figure) has a much higher average negative log likelihood according to the input document than according to the summary.
	}
	\label{fig:log-likelihood-histogram}%
\end{figure}

We take an example to illustrate how \textsc{Quals} captures the factual inconsistency of summaries. Below is a generated summary from the BART-large MLE model on a test example: 
\begin{quotation}
	\emph{The AirAsia flight \textcolor{red}{4U 9525} crash was the latest in a series of tragedies that have hit the aviation industry.}
\end{quotation}
The input document described the AirAsia crash but did not mention the flight number. In fact, ``4U 9525'' is the Germanwings flight that crashed in the French Alps. The model hallucinated the flight number because it appeared in several training examples that cover the Germanwings crash. Given the above summary, our QAGen model generates the following q-a pairs: \emph{Q1: What was the name of the flight that crashed? A1: 4U 9525. Q2: Which airlines flight crashed? A2: AirAsia}. In Figure \ref{fig:log-likelihood-histogram} we show the negative log likelihood per subword token on these q-a pairs conditioned on the summary (blue) and input document (orange). The answer to the first question is very likely according to the summary while extremely unlikely according to the input document, indicating factual inconsistency. On the other hand, ``AirAsia'' is factually consistent and the second q-a pair is as likely according to the summary as the input document. The \textsc{Quals} score for the two q-a pairs are $-2.615$ and $-0.054$, respectively.

\subsection{More qualitative examples}
We provide additional qualitative examples in Table \ref{tab:more_qualitative_analysis} for summaries generated by MLE and \conseq. In Table \ref{tab:more_qualitative_qagen}, we show additional examples of the questions and answers generated by QAGen model based on the summaries. 
\begin{table*}[]
	\centering
	\resizebox{1.0\textwidth}{!}{%
		\scriptsize
		\begin{tabular}{p{0.1\linewidth}p{0.8\linewidth}}
			\toprule
			Input 1: & ...Eventually she decided to train for the 10,000m and her win at Parliament Hill, 17 seconds inside the GB qualification time, came in only her second track event at the distance...."I haven't been racing on the British scene recently, so that's why it's come as quite a shock."..."We'll start to plan the build-up, the training and the races I'll compete in before Rio." \\
			MLE: & British 10,000m runner Sophie Andrews says she has been "blown away" by the reaction to her shock victory at the \textcolor{red}{London Anniversary Games}. \\
			\conseq: & British 10,000m runner Sophie Andrews says she is still in shock after winning at Parliament Hill in London to qualify for the Rio Olympics.
			\\  \cmidrule(l){1-2}
			Input 2: & ... The Plymouth University marine biology student said she only realised the cosmetic prosthetic was missing the following morning. She thinks she could have left it in a nightclub and forgot to pick it up.... \\
			MLE: & A university student who lost her hand in a nightclub has said she is ``devastated'' to have lost it.
			\\ 
			\conseq: & A student who lost her prosthetic hand in a nightclub has said it is ``worth everything'' to her.
			\\ \cmidrule(l){1-2}
			Input 3: & The SNP topped the vote in the Leith Walk by-election, while Scottish Labour won the second seat from the Greens. ... The SNP's John Lewis Ritchie topped the Leith Walk poll with 2,290 votes. He was elected at stage one in the STV process with a swing in first-preference votes of 7.6\% from Labour. ... \\
			MLE: & The SNP has won the Leith Walk by-election and the \textcolor{red}{Scottish Greens have won the Midlothian West seat}. \\
			\conseq: & The SNP has won the Leith Walk by-election in Edinburgh with a swing of 7.6\% from Labour. \\ \cmidrule(l){1-2}
            Input 4: & Conrad Clitheroe, left, and Gary Cooper, right, were thrown in jail after being arrested for writing down aircraft registration numbers in Dubai. Three British men who have spent two months in prisons in Dubai and Abu Dhabi after they were seen plane spotting are to be freed. Conrad Clitheroe, 54 and Gary Cooper, 45, from Stockport, along with their ex-pat friend Neil Munro were stopped by police for writing down aircraft registration numbers at Fujairah Airport. They were taken to a police station and despite being told they would not be detained, were put into prison. ... \\
            MLE: & Conrad Clitheroe, Gary Cooper and Neil Munro were stopped at Fujairah Airport. They were taken to a police station and told they would not be detained. But \textcolor{red}{despite being told they would be detained}, they were put into prison. Families were concerned for their welfare as Mr Clitheroe has a heart condition. Today they were told they would be able to go home after the court accepted a defense argument they were pursuing their passion for plane spotting. \\
            \conseq: & Conrad Clitheroe, Gary Cooper and Neil Munro were stopped by police in Dubai. They were taken to a police station and despite being told they would not be detained, were put into prison. Families of the three men confirmed they would finally be able to return home and charges of espionage would not be brought against them. Mr Clitheroe, who suffers from a heart condition, will be able to celebrate his first wedding anniversary with his wife Valerie on May 9. \\ \cmidrule(l){1-2}
            Input 5: & ..Now police investigating the case have come across CCTV that shows the man following three other potential victims earlier on the same night. ... It includes one woman who was forced to flee from the man by bus after walking to her car in Leeds city centre, ending up miles away from where she was going, just to get to safety. Another potential victim followed may have been aware she was being stalked as he also walked up behind her in the city centre as she walked past a row of bars. While a third woman can be seen on the footage seen by police turning around to find herself just an inch away from the rapist, before darting into a shop and staying in there until he left on the city's Burley Road. ... He told the Daily Mirror: 'His first target was a woman in her 20s who had parked her car. He is seen walking behind her, six inches from her back pocket. 'She was so concerned she got a bus that took her miles from where she was going.\\
            MLE: & Woman, 18, was attacked at a bus stop in Beeston, Leeds, on March 6. She was hit over the head 20 times with a rock and left for dead. Police have now found CCTV showing the man following three other women. One woman was forced to flee by bus to get to safety from the rapist. Another woman \textcolor{red}{in her 20s was just an inch away from the rapist}. The man is wanted for attempted murder and rape after the 'appalling' attack. \\
            \conseq: & The 18-year-old was hit over the head 20 times with a rock and dragged into a garden to be brutally raped by the man and left for dead. Police investigating the case have come across CCTV that shows the man following three other potential victims earlier on the same night. One woman was forced to flee from the man by bus after walking to her car, ending up miles away from where she was going. Another woman can be seen turning around to find herself just an inch away from the rapist, before darting into a shop. \\ \cmidrule(l){1-2}
            Input 6: & ...A new metro line extension opened this week in the Czech capital, transporting travellers closer to the airport than ever before‚ but the point where passengers transfer from trains to buses doesn't have a moving staircase. ... \\
            MLE: & New metro line extension opened this week but there is no moving staircase. That means porters are left to carry heavy bags up and down 32 stairs. One of the porters said he wanted to 'stay fit and help the elderly' Officials are now considering adding an escalator while the train project is completed. \\
            \conseq: & A new metro line extension opened this week in the Czech capital. But the point where passengers transfer from trains to buses doesn't have a moving staircase. That means porters are left to carry heavy bags up and down 32 stairs. Officials are now considering adding an escalator while it remains unclear when the train project will be completed.\\ \cmidrule(l){1-2}
			Input 7: & At least two have been treated for symptoms of poisoning after being walked near a cricket pitch in Fordham Heath ... The dogs which were treated, both suffered "severe sickness" after being walked in a field near Eight Ash Green Cricket Club in Spring Lane ... \\
			MLE: & A number of dogs have been taken ill after being walked in a field near a cricket club. \\
			\conseq: & Two dogs have suffered "severe sickness" after being walked in a field near a cricket club. \\
			\bottomrule
		\end{tabular}%
	}
	\caption{MLE summaries in Example 1  contains hallucination. MLE summary in Example 2 was misleading by omitting ``prosthetic''. MLE summary in Example 3 contradicted the input document, which says Scottish Greens lost. MLE summary in Example 4 misses a ``not''. MLE summary in Example 5 confuses the victims; the woman in her 20s fled by bus according to the input document. The \conseq\ summary in Example 6 is more informative as it mentions the metro line is located in the Czech capital. Example 7: input document mentioned ``\emph{at least} two dogs'' but \conseq\ omitted ``\emph{at least}''.
	}
	\label{tab:more_qualitative_analysis}
\end{table*}

\begin{table*}[]
	\centering
	\resizebox{1.0\textwidth}{!}{%
		\scriptsize
		\begin{tabular}{p{0.1\linewidth}p{0.8\linewidth}}
			\toprule
			Summary1: & A security alert at the Alliance Party offices in Belfast city centre is being considered a hoax, its chairman has said. \\
			Q\&A: & What is the security alert considered? a hoax \\
			Q\&A: & In what city is the security alert located? Belfast \\
			Q\&A: & Who said that the security alert was a hoax? chairman \\
			Q\&A: & What is the security alert considered? a hoax \\
			Q\&A: & What party is the security alert at? Alliance Party \\
			Q\&A: & What is considered a hoax? security alert at the Alliance Party offices in Belfast city centre \\ \cmidrule(l){1-2}
			Summary2: & The UK's international reputation has been knocked, with a global energy watchdog warning the UK faces a loss of status as a leader in clean energy. \\
			Q\&A: & What is the name of the global energy watchdog? Global energy watchdog \\
			Q\&A: & What has been knocked? The UK's international reputation \\
			Q\&A: & Which country's reputation has been knocked? The UK \\
			Q\&A: & What does the global energy watchdog say the UK faces? loss of status as a leader in clean energy \\
			Q\&A: & Who warned that the UK could lose its clean energy leadership status? a global energy watchdog \\
			Q\&A: & In what field is the UK losing its status as a leader? clean energy \\
			Q\&A: & what did global energy watchdog say? the UK faces a loss of status as a leader in clean energy. \\
			Q\&A: & In what field is the UK losing its status as a leader? clean energy. \\
			Q\&A: & How much of a reputation has the UK lost in the world? The UK's international reputation has been knocked, \\
			Q\&A: & In what field is the UK losing its status as a leader? clean energy \\
			Q\&A: & Is the UK's international reputation good or bad? has been knocked \\
			Q\&A: & What kind of reputation has been knocked? international \\
			Q\&A: & What kind of reputation has been knocked by the global energy watchdog? international reputation \\
			Q\&A: & What does the global energy watchdog say the UK faces? loss of status as a Leader in clean energy. \\ \cmidrule(l){1-2}
			Summary3: & Former Italy and South Africa coach Nick Mallett says southern hemisphere sides need to put themselves in the same category as their World Cup rivals. \\
			Q\&A: & What does Nick Mallett say? southern hemisphere sides need to put themselves in the same category as their World Cup rivals \\
			Q\&A: & What team does Nick Mallett coach? Italy and South Africa \\
			Q\&A: & Which coach says that southern hemisphere teams need to put themselves in the same category as their World Cup rivals? Nick Mallett \\
			Q\&A: & What did Mallett say? southern hemisphere sides need to put themselves in the same category as their World Cup rivals. \\
			Q\&A: & What type of competition does Nick Mallett coach? southern hemisphere sides \\
			Q\&A: & Along with South Africa, what country does Nick Mallett coach? Italy \\
			Q\&A: & Whose name does Nick Mallett use as a coach? Italy and South Africa coach \\
			Q\&A: & Who says that southern hemisphere teams need to put themselves in the same category as their World Cup rivals? Nick Mallet \\
			Q\&A: & Nick Mallett is the coach of what team? Italy and South Africa \\
			Q\&A: & Name the coach of Italy and South Africa? Nick Mallett \\
			Q\&A: & What does Nick Mallett say? southern hemisphere sides need to put themselves in the same category as their World Cup rivals. \\
			Q\&A: & To whom did Nick Mallett compare the world cup rivals? southern hemisphere sides \\
			Q\&A: & Where does Nick Mallett work? Italy and South Africa coach \\
			Q\&A: & Which coach says that southern hemisphere teams need to put themselves in the same category as their World Cup rivals? Nick Mallet \\ \cmidrule(l){1-2}
			Summary4: & New Bury boss Lee Clark has confirmed the appointment of former Sunderland striker Rob Wilson as first-team manager on a two-year contract. \\
			Q\&A: & Who is the Sunderland striker? Rob Wilson \\
			Q\&A: & What team did Rob Wilson play for? Sunderland \\
			Q\&A: & How long is the contract that Rob Wilson has? two-year \\
			Q\&A: & who appointed Rob Wilson? Lee Clark \\
			Q\&A: & What team did Rob Wilson play for? Sunderland striker \\
			Q\&A: & When did Lee Clark confirm the appointment of Rob Wilson? two-year contract \\
			Q\&A: & On what kind of contract did Rob Wilson sign? first-team manager \\
			Q\&A: & Who is the new manager of the Bury football team? Rob Wilson \\
			Q\&A: & what job has Lee Clark done? Bury boss \\
			Q\&A: & Rob Wilson is a former Sunderland striker for what team? Bury \\ \cmidrule(l){1-2}
			Summary5: & An image thought to be by street artist Banksy appeared on a shop wall on Sunday during the Queen's Jubilee. \\
			Q\&A: & When did Banksy's image appear on a shop wall? Sunday \\
			Q\&A: & What type of building has been built over a tree? building \\
			Q\&A: & Which artist was thought to be Banksy? street artist \\
			Q\&A: & What was the name of the artist who was on the wall? Banksy \\
			Q\&A: & Where did Banksy's image appear? on a shop wall \\
			Q\&A: & Which artist was thought to be Banksy? street artist Banksy \\
			Q\&A: & A Banksy image appeared on a shop wall during what event? Queen's Jubilee \\
			Q\&A: & On what day did Banksy's image appear on a shop wall? Sunday \\
			Q\&A: & when the image appeared? Sunday during the Queen's Jubilee \\
			Q\&A: & What event did Banksy appear on a shop wall during? Queen's Jubilee. \\
			Q\&A: & On what day did Banksy's image appear on a shop wall? Sunday \\
			Q\&A: & Banksy's image was thought to be by who? street artist \\
			Q\&A: & Who painted an image on Sunday? Banksy \\
			Q\&A: & What was the name of the artist who was on the wall? Street artist Banksy \\
			Q\&A: & What event did Banksy appear on a shop wall during? Queen's Jubilee \\
			Q\&A: & When did this happen? Sunday during the Queen's Jubilee \\
			\bottomrule
		\end{tabular}%
	}
	\caption{Additional examples of the question and answers generated by the QAGen model based on the summaries.
	}
	\label{tab:more_qualitative_qagen}
\end{table*}
\subsection{Human evaluation}
\begin{figure*}%
	\centering
	\includegraphics[scale=0.35]{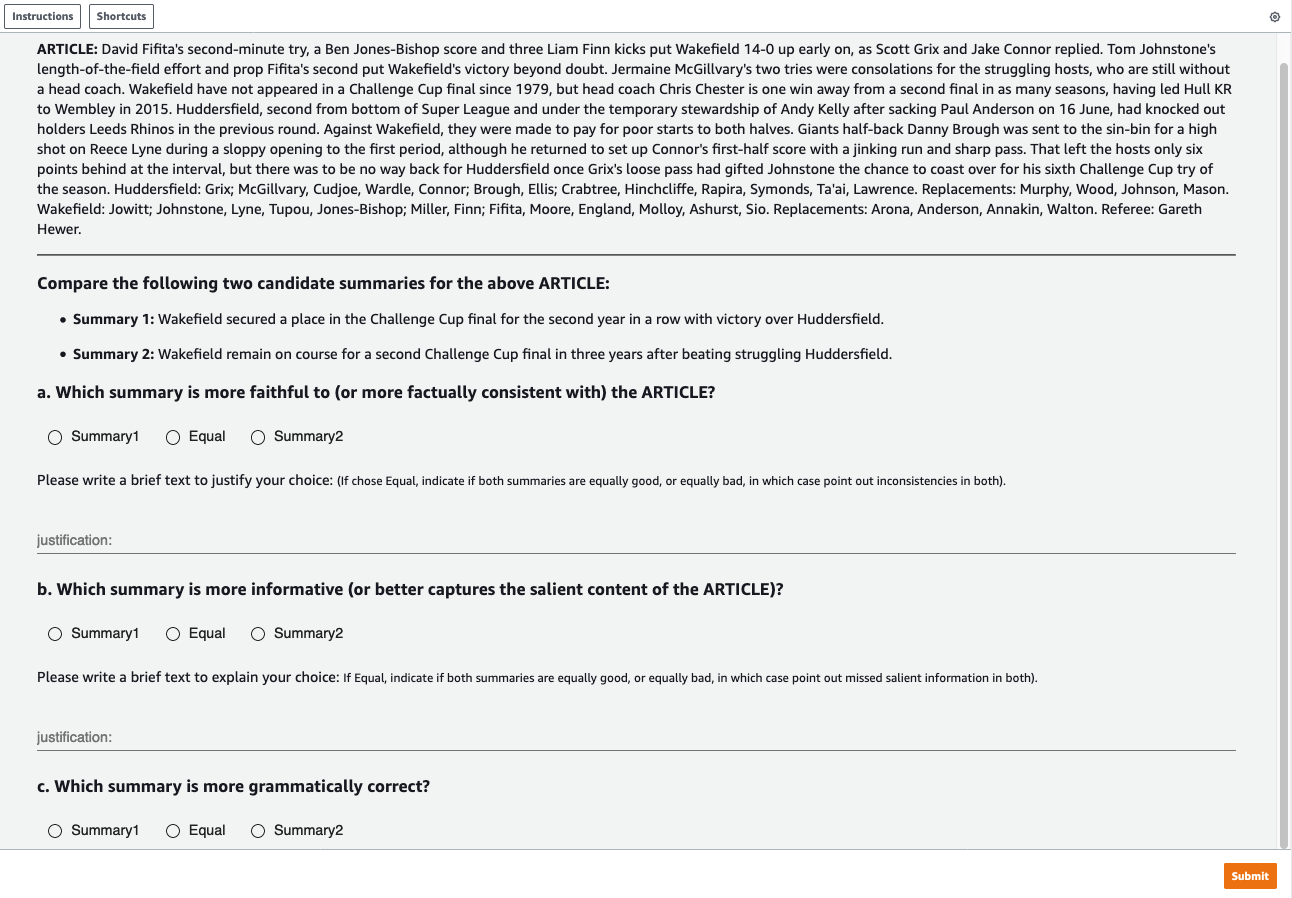}
	\caption{Human evaluation interface using Amazon Sagemaker Ground Truth.}
	\label{fig:human-eval-screenshot}%
\end{figure*}

\begin{figure*}%
	\centering
	\subfigure{%
		\includegraphics[width=0.7\textwidth,height=0.3\textheight]{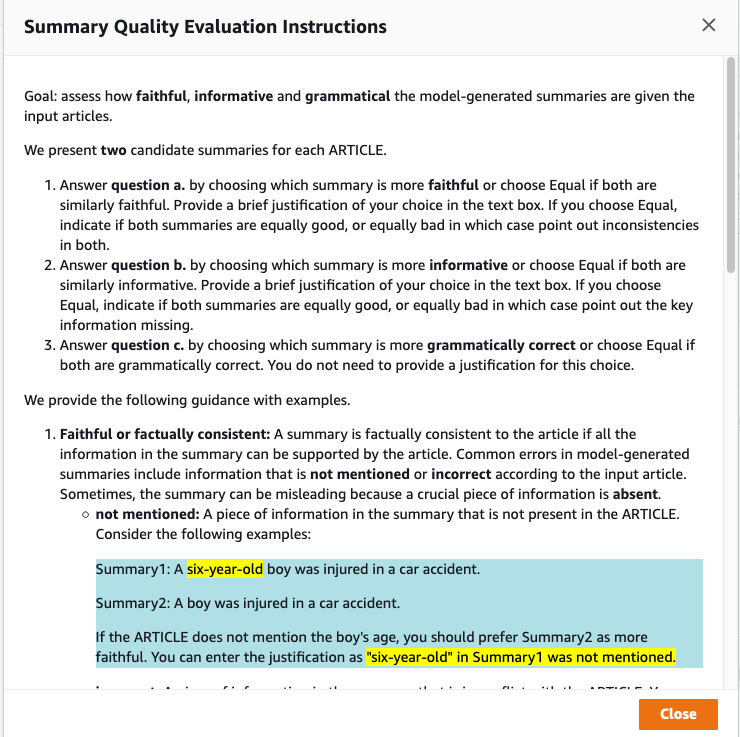}}\\
	\subfigure{%
		\includegraphics[width=0.7\textwidth,height=0.3\textheight]{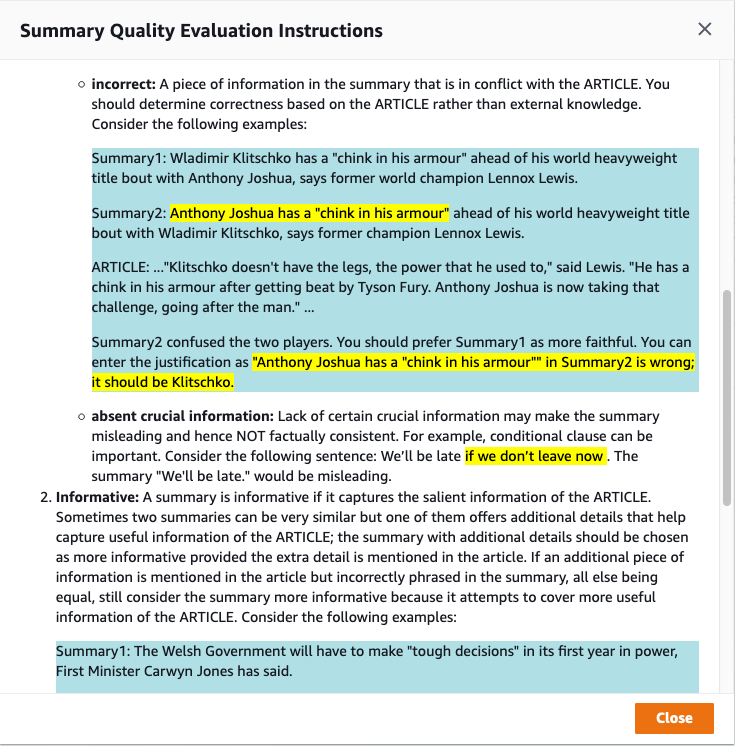}}\\
	\subfigure{%
		\includegraphics[width=0.7\textwidth,height=0.3\textheight]{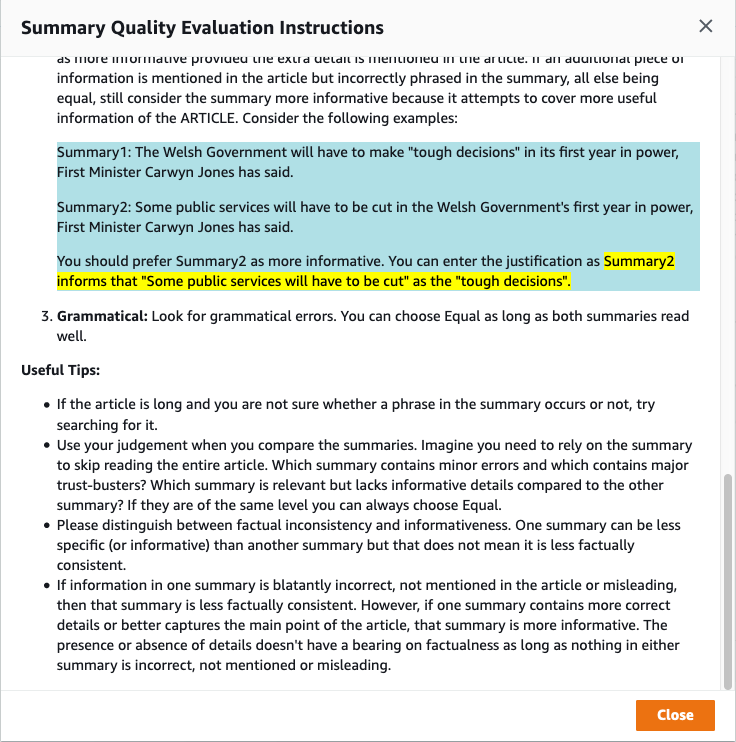}}
	\caption{Human evaluation instruction screenshots.
	}
	\label{fig:human-eval-instructions}%
\end{figure*}

We show the interface of our human evaluation in the Figure \ref{fig:human-eval-screenshot}. The full instruction for the annotators are shown in Figure \ref{fig:human-eval-instructions}

%

\end{document}